\def\year{2022}\relax
\documentclass[letterpaper]{article} 
\usepackage{aaai22}  
\usepackage{times}  
\usepackage{helvet}  
\usepackage{courier}  
\usepackage[hyphens]{url}  
\usepackage{graphicx} 
\usepackage{natbib}  
\usepackage{caption} 
\usepackage{algorithm}
\usepackage{algorithmicx}
\usepackage{algpseudocode}
\DeclareCaptionStyle{ruled}{labelfont=normalfont,labelsep=colon,strut=off} 
\usepackage{newfloat}
\usepackage{listings}
\floatstyle{ruled}
\newfloat{listing}{tb}{lst}{}
\floatname{listing}{Listing}
\usepackage{multirow}
\usepackage{pifont}
\usepackage{amsmath, amsfonts}
\usepackage{algorithm}
\usepackage{algorithmicx}
\usepackage{algpseudocode}
\usepackage{subfigure}

\newcommand{\Ours}{SentiPrompt}
\title{SentiPrompt: Sentiment Knowledge Enhanced Prompt-Tuning for\\ Aspect-Based Sentiment Analysis}

\author{
    Chengxi Li\textsuperscript{\rm 1,2}, 
    Feiyu Gao\textsuperscript{\rm 2},
    Jiajun Bu\textsuperscript{\rm 1,3}, 
    Lu Xu\textsuperscript{\rm 2},
    Xiang Chen\textsuperscript{\rm 1}, 
    Yu Gu\textsuperscript{\rm 1},
    Zirui Shao\textsuperscript{\rm 1},\\
    Qi Zheng\textsuperscript{\rm 2},
    Ningyu Zhang\textsuperscript{\rm 1,3},
    Yongpan Wang\textsuperscript{\rm 2},
    Zhi Yu\textsuperscript{\rm 1,3}\footnote{Corresponding author} \\
}
\affiliations{
    \textsuperscript{\rm 1}Alibaba-Zhejiang University Joint Institute of Frontier Technologies, Hangzhou,China \\
    \{chengxili,bjj,xiang\_chen,zhangningyu,yuzhirenzhe\}@zju.edu.cn \\
    \textsuperscript{\rm 2}Damo Research, Alibaba Group\\
    \{feiyu.gfy,lu.x,yongqi.zq\}@alibaba\--inc.com \\
    \textsuperscript{\rm 3}Ningbo Research Institute, Zhejiang University, Ningbo, China \\
}

\usepackage{bibentry}
\begin{document}

\maketitle

\begin{abstract}
\textit{Aspect-based sentiment analysis} (ABSA) is an emerging fine-grained sentiment analysis task that aims to extract aspects, classify corresponding sentiment polarities and find opinions as the causes of sentiment. The latest research tends to solve the ABSA task in a unified way with end-to-end frameworks. Yet, these frameworks get fine-tuned from downstream tasks without any task-adaptive modification. Specifically, they do not use task-related knowledge well or explicitly model relations between aspect and opinion terms, hindering them from better performance. In this paper, we propose \Ours~to use sentiment knowledge enhanced prompts to tune the language model in the unified framework. We inject sentiment knowledge regarding aspects, opinions, and polarities into prompt and explicitly model term relations via constructing consistency and polarity judgment templates from the ground truth triplets. Experimental results demonstrate that our approach can outperform strong baselines on Triplet Extraction, Pair Extraction, and Aspect Term Extraction with Sentiment Classification by a notable margin. 
\end{abstract}

\section{Introduction}
\begin{figure}[htb]
    \centering
    \includegraphics[width=0.49\textwidth,]{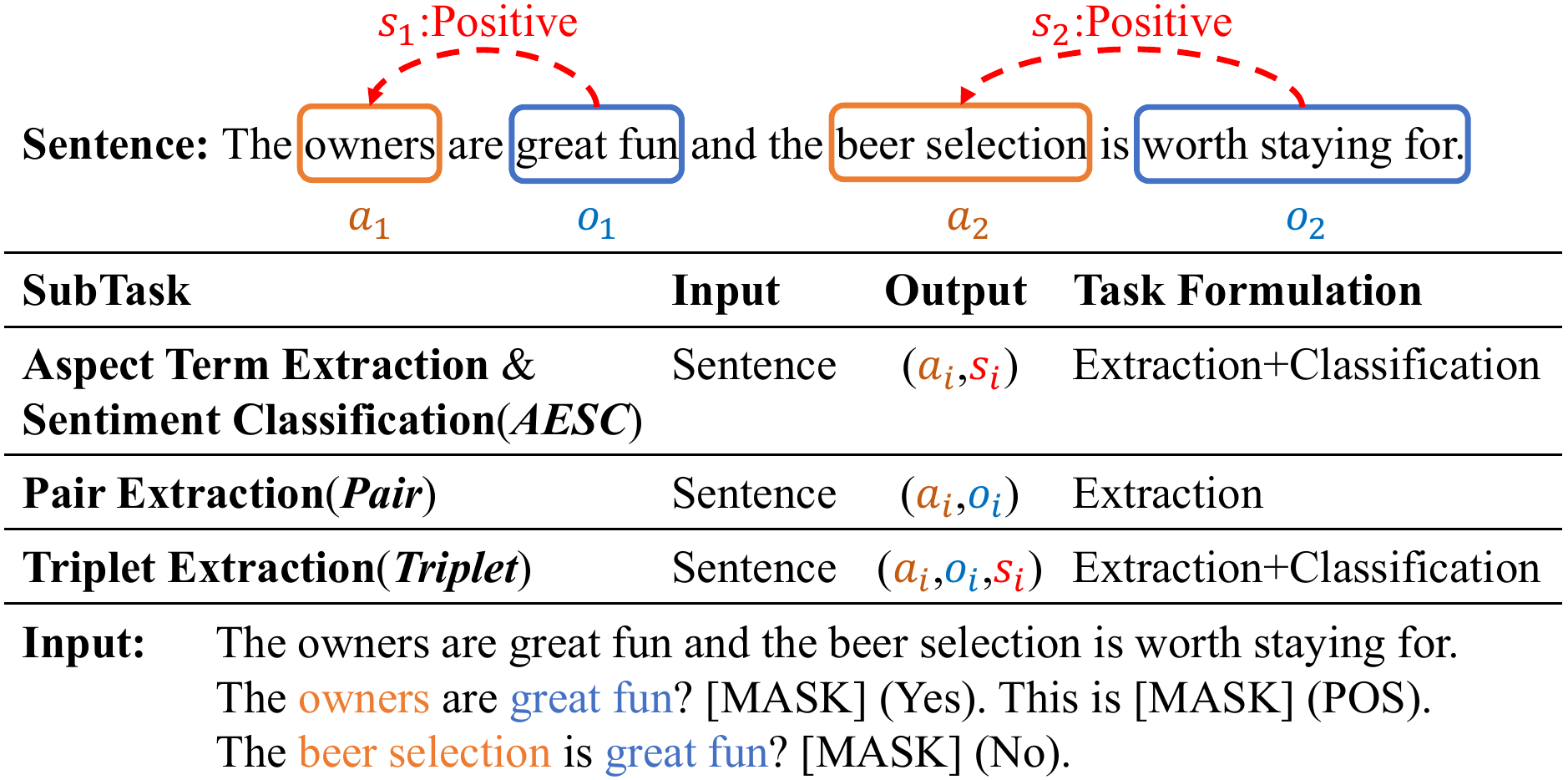}
    \caption{Examples of prompt-tuning for ABSA subtasks.}
    \label{fig:subtask_example}
\end{figure}

Aspect-based Sentiment Analysis~\citep{pontiki-etal-2014-semeval} is a
fine-grained sentiment analysis task which requires extracting aspects, judging the corresponding polarities, and finding opinions as the causes of sentiment towards each aspect. 
For the example in Figure \ref{fig:subtask_example}, the task aims to extract the two aspects ``owners'' and ``beer selection'', their corresponding opinions (``great fun'' and ``worth staying for'') and polarities (both positive). 
Previous studies have proposed many subtasks for ABSA, all being different in required input, output and task formulations. 
In this paper, we focus on the most complex and challenging three subtasks, \emph{ASEC}, \emph{Pair}, and \emph{Triplet}, as shown in Figure \ref{fig:subtask_example}.
Aspect Term Extraction and Sentiment Classification (\emph{AESC}) requires extracting the aspect terms and classifying the sentiment polarities about them.
Pair Extraction (\emph{Pair}) extracts the aspect
terms as well as the corresponding opinion terms simultaneously. 
Aspect Sentiment Triplet Extraction (\emph{Triplet})~\citep{Peng_2020} aims to get the information containing all the aspects, their corresponding opinions and sentiment polarities from input sentences. These are all nearly complete solutions for the ABSA task, especially \emph{Triplet}.

Researchers have proposed pipeline approaches~\citep{Peng_2020,Mao2021AJT} to solve these subtasks in a multi-stage framework. 
However, these multi-stage models are not end-to-end, thus, making predictions from various sub-models inside separately.
Separating tasks into multiple stages potentially breaks the relation modeling within the triplets and brings about ineligible error propagation. 
Recently, several neural-network-based models~\citep{wu-etal-2020-grid,xu-etal-2020-position,xu-etal-2021-learning} have been proposed to develop end-to-end framework with sequence tagging. 
In these approaches, aspects and opinions can be jointly extracted and polarities are also jointly considered.
However, these approaches mainly rely on modeling the relations between aspects and opinions at the word level, regardless of the span-level relations which need supervisions from extra subtasks for pruning.
More recent studies make attempts on unified frameworks with Pre-trained Langugage Models (PLM) for the ABSA task~\citep{Chen2021BidirectionalMR,yan2021unified}, yielding promising performance on benchmark datasets. 
Despite the success of PLM fine-tuning, some recent studies find that one of its critical challenges is the significant gap of objective forms between pre-training and fine-tuning, which restricts PLMs from reaching their full potential.

Prompt-tuning~\citep{Jiang_2020,autoprompt,PET2021,PTR2021}, as a new fine-tuning paradigm, has emerged which is proposed to bridge the gap of objective forms between pre-training and fine-tuning. With appropriate prompts and tuning objectives, the prompt-tuning could manipulate the model behavior to fit various downstream tasks.
By using specially constructed prompts, we can further inject and stimulate the task-related knowledge in PLMs, thus boosting the model performance.
Nevertheless, handcrafting the appropriate prompt for ABSA requires domain expertise, and auto-constructing a well-performing prompt often requires additional computational cost for verification. 


Motivated by the above observations, in this paper, we propose \Ours, a novel approach to leverage sentiment knowledge enhanced prompts to tune the LM in a generative framework for ABSA.
To be specific, we leverage consistency and polarity judgment templates to construct prompts regarding ground truth aspects, opinions, and polarities as sentiment knowledge. 
During training, the model predicts the consistency of prompt samples with the ground truth and classify polarities for consistent ones as instructed. In this way, the encoder part of the framework learn to explicitly capture sentiment relations between aspect and opinion term pairs. Similarly, we introduce sentiment knowledge by this tuning process. 
Together with the supervision from generative predictions, the framework can serve the ABSA task far better than solely being fine-tuned by the loss from final predictions.
This is the first attempt to introduce downstream task-related knowledge and to enhance relation modeling for the LM through prompt-tuning methods.
Sentiment knowledge enhanced prompt-tuning, working as an ABSA task-adaptive modification to the tuning process, helps to promote term extraction, pairing and classification with sentiment knowledge enhanced prompts.
This technique needs no extra parser to get semantic information or building graph from sentence structure. 
Besides, neither aspect and opinion terms nor polarities are of large numbers in a comment. 
Thus it allows convenient, time-sparing and explicit modeling of relations between those terms and polarities and better introduction of sentiment-related knowledge for the LM. 

Our main contributions can be summarized as follows:
\begin{itemize}
    \item We propose sentiment knowledge enhanced prompt-tuning (\Ours) to explicitly model the relations between terms and leverage sentiment knowledge. 
    We further utilize trainable continuous templates to obtain optimized prompts. 
    To the best of our knowledge, this is the first approach to incorporate prompt-tuning for ABSA subtasks.
    \item We revise the latest version of four benchmark datasets regarding mistakes and incompleteness in triplet annotations and release a new version for future research. Experimental results demonstrate that our method archives promising improvements and obtains new state-of-the-art performance all datasets.
\end{itemize}
\section{Related Works}
\subsection{Aspect-based Sentiment Analysis}
Early research about the ABSA task starts with extracting aspects or opinions alone.
Most early research follows sequence tagging schema~\citep{ijcai2018-583ae,xu-etal-2020-position,bi2021interventional} to solve Aspect Term Extraction (\emph{ATE}) while recent works~\citep{ma-etal-2019-exploring-seq-2-seq-ae,li-etal-2020-conditional-seq2seq-ae} attempt to use sequence-to-sequence techniques with PLMs and achieve relatively promising results. As for Opinion Term Extraction (\emph{OTE}), most works treat it as an auxiliary task~\citep{wang-pan-2018-recursive, he-etal-2019-interactive,chen-qian-2020-relation}.
Later research turns to extract the corresponding opinions (Aspect-oriented Opinion Extraction, \emph{AOE}) or classifying polarities (Aspect-level Sentiment Classification, \emph{ALSC}) for every given aspect.
\citet{fan-etal-2019-target} first introduces \emph{AOE} with their baseline model and datasets. Lately, others~\citep{Wu_Zhao_Dai_Huang_Chen_2020,pouran-ben-veyseh-etal-2020-introducing} develop some sequence tagging models to follow this track. For \emph{ALSC}, works about it start earlier. Most of them~\citep{tang-etal-2016-effective,wang-etal-2016-attention} incorporate long-short-term-memory(LSTM) units or attention mechanism to deal with relations between aspects and sentiments. In addition, some works~\citep{chen-etal-2020-inducing,wang-etal-2020-relational,tang-etal-2020-dependency,li-etal-2021-dual-graph} propose to leverage graph neural networks to utilize syntax and semantic information from extra parsers.

Recent studies propose some more fine-grained subtasks \emph{AESC}, \emph{Pair} and \emph{Triplet} for ABSA.
\citet{Peng_2020} first proposes a multi-stage pipeline to do extraction, pairing and classification step by step.
Then \citet{xu-etal-2020-position} follows the commonly used sequence tagging schema with a position-aware method.
Yet another line~\citep{wu-etal-2020-grid} tries to solve \emph{Triplet} by labeling the relations of word-pairs.
\citet{xu-etal-2021-learning} goes further by labeling span-pairs with a pruning method.
Moreover, \citet{Mao2021AJT,yan2021unified} propose to combine a set of subtasks through unified formulations. They formulate various subtasks as a machine reading comprehension or generation task, which perform quite well.
Our method tends to solve this task in a unified generative framework yet include explicit modeling of relations between terms through sentiment knowledge enhanced prompt-tuning.

\subsection{Prompt-tuning}
Prompt-tuning is a new paradigm of fine-tuning inspired by GPT-3~\citep{gpt3}, especially for language models in few-shot or zero-shot settings.
It means prepending instructions and demonstrations to the input and output predictions.
Recent PET work~\citep{PET2021} focus on semi-supervised setting with a large number of unlabeled examples.
\citet{Gao2020MakingPL} explore their prompt-tuning methods with demonstrations on language models for some benchmark tasks, including Sentiment Classification.
Prompt-tuning can induce better performances for PLMs on widespread of NLP tasks including text classification \cite{Gao2020MakingPL,zhang2021differentiable}, relation extraction \cite{PTR2021,chen2021adaprompt}, NER \cite{chen2021lightner}, entity typing \cite{ding2021prompt}, and so on.
To construct better prompt for downstream tasks, several approaches \cite{PTR2021,hu2021knowledgeable} leverage knowledge injection \cite{zhang2021drop,zhang2021alicg} to templates and vertalizer construction.
Besides, there exist lots of works on prompting for mining knowledge from PLMs~\citep{Davison_2019,Petroni_2019,Talmor_2020}.
Unlike those works, we focus on ABSA which is quite different from classification tasks.
Furthermore, since handcrafting a well-performing prompt is rather difficult, a number of works focus on automatic prompt search via generation or gradient-based search ~\citep{Jiang_2020,autoprompt,Gao2020MakingPL}.
Those works mentioned above still keep the discrete format for prompt search whereas \citet{liu2021gpt,Li2021PrefixTuningOC} tends to do optimization for continuous prompt embeddings.

\section{Methodology}
In this section, we first present the task formulation and the sequential output generated to cover all subtasks, then elaborate our proposed method about sentiment knowledge enhanced prompt-tuning and the generation module. 
\subsection{Task Formulation}
\label{sec:formulation}
For a given sentence $X$, we tokenize it into a sequence of tokens $X=[x_1, x_2, ...x_n]$. The ABSA task aims to provide the aspect terms, the sentiment polarity, and the opinion terms, represented by \emph{a}, \emph{s} and \emph{o} in our method respectively. \emph{a} and \emph{o} are span indexes of terms and $\emph{s}\in \{\mbox{POS}, \mbox{NEG}, \mbox{NEU}\}$ is the polarity class index. Superscripts \textsuperscript{start} and \textsuperscript{end} denote the start and end index of the corresponding token span in the sequence. For our generative framework, the target sequence $Y = [y_1, y_2, ...., y_m]$ consists of multiple base prediction $pred = [a^{start}, a^{end}, o^{start}, o^{end}, s]$. Base predictions of subtasks are all some subsets of the above $pred$ and we list them as follows: 
\begin{itemize}
    \item For \emph{AESC}: $pred = [a^{start}, a^{end}, s]$
    \item For \emph{Pair}: $pred = [a^{start}, a^{end}, o^{start}, o^{end}]$
    \item For \emph{Triplet}: $pred = [a^{start}, a^{end}, o^{start}, o^{end},s]$
\end{itemize}
In our work, we use BART~\citep{Lewis_2020_bart} with pointer network~\citep{NIPS2015_pointer} for sequence-to-sequence generation. 
During training, the input $X$ is used for both sentiment knowledge enhanced prompt-tuning and index sequence generation. The encoder part is shared across both sides.

\begin{figure*}[htb]
    \centering

    \subfigure[\Ours~Tuning]{
        \includegraphics[width=0.49\textwidth,]{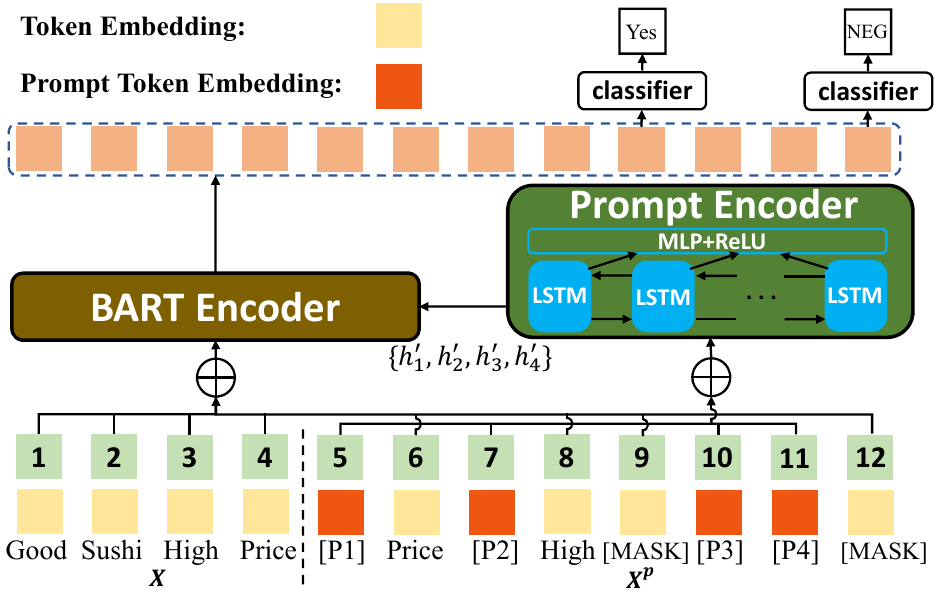}
    }
    \hspace{-.1in}
    \subfigure[Generation Framework]{
        \includegraphics[width=0.49\textwidth,]{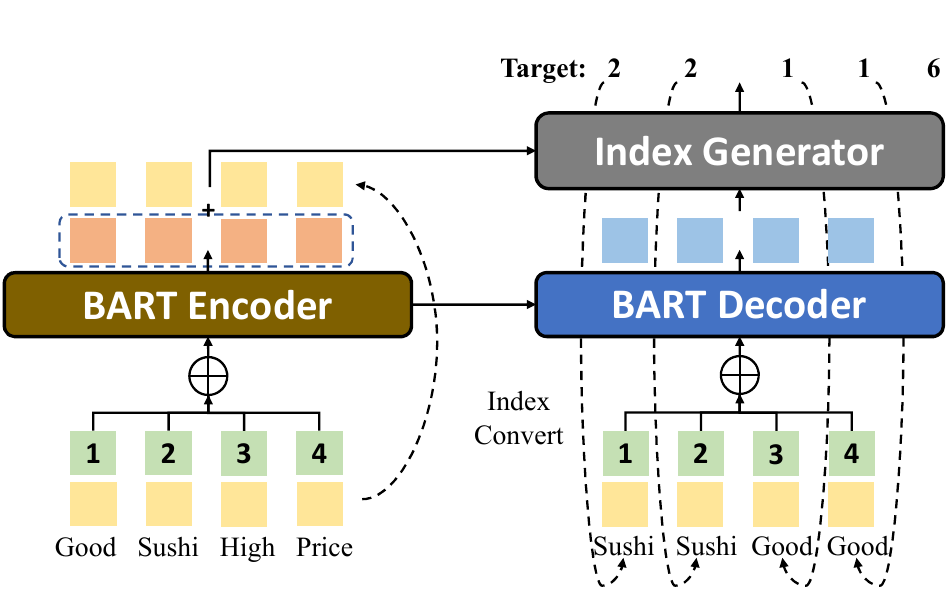}
    }

    \caption{The overall architecture of our model. 
    Embedding vectors represented as light yellow and orange boxes are retrieved from different embedding layers (Best viewed in color).}
    \label{fig:model}
\end{figure*}

\subsection{\Ours~Tuning}
\begin{table*}[htb]
\small
    \centering
    \begin{tabular}{c|c|c}
    \hline
         & Manually Designed & Auto-constructed  \\
    \hline
         \multirow{2}{*}{Consistency Prompt} & The \textit{Sushi} is \textit{High}? [MASK]
         & $P_1$,$P_2$..$P_{l_1}$ \textit{Sushi} $P_{l_1+1}$,$P_{l_1+2}$...$P_{l_2}$ \textit{High}? [MASK]\\
         & The \textit{Good Sushi} is \textit{High}? [MASK]
         & $P_1$,$P_2$...$P_{l_1}$ \textit{Good Sushi} $P_{l_1+1}$,$P_{l_1+2}$...$P_{l_2}$ \textit{High}? [MASK]\\
    \hline
         Polarity Prompt & This is [MASK]
         & $P_{l_2+1}$,$P_{l_2+2}$...$P_{l_P}$,[MASK]\\
    \hline
    \end{tabular}
    \caption{The demonstration of our prompt designs taking ``\textit{Good Sushi High Price.}'' as the input example.}
    \label{tab:prompt_example}
\end{table*}
The sentiment knowledge of ground truth sentiment terms and polarities is injected in the construction of prompts. The prompt encoder is inserted here to provide trainable and continuous representations for prompts.

\subsubsection{Sentiment Knowledge Enhanced Prompt Construction}
\Ours~tunes the LM with sentiment knowledge enhanced prompts as a masked language modeling task (MLM) during training.
So we construct prompts as samples for \Ours~MLM based on terms and polarities from ground truth triplets. To construct a prompt masked training sample, we randomly take an aspect and an opinion from ground truth triplets to fill in a fixed template. 
Given an input sequence $X$ (e.g. \textit{Good Sushi High Price}), a template $T$ consists of pseudo prompt tokens $P_k$ and randomly sampled aspect and opinion terms $A$, $O$ (sampled from (\textit{Sushi}, \textit{Price}) and (\textit{Good}, \textit{High})). Here $P_k$ can be specific English words or just a randomly-initialized pseudo token to be optimized. It is formed like:
\begin{equation}
    [P_{1:l_1},A,P_{l_1+1:l_2},O,\mbox{[MASK]}]\\
\end{equation}
Besides, we randomly do span manipulations to aspects and opinions of some prompts in construction. These prompts will be constructed with a sub-span of aspect and opinion terms if these terms contain more than one token. For single token cases, we attach a few neighbor tokens of the term to it. 
On the whole, the created prompts may be consistent with the ground truth triplets or against them due to random sample choices and random span manipulations. For the example input, the ground truth triplets are (\textit{Sushi}, \textit{Good}, POS) and (\textit{Price}, \textit{High}, NEG). Therefore, other possible pairs are all considered inconsistent. In Table \ref{tab:prompt_example}, the first example is constructed from (\textit{Sushi}, \textit{High}) and the second by attaching the left neighbor \textit{Good} to the aspect \textit{Sushi}.
For chosen pairs that are consistent with the ground truth, we will attach an extra polarity judgement prompt to it as the suffix. 
Likewise, the template $T'$ for the polarity judgement is formed like:
\begin{equation}
    T' = [P_{l_2+1:l_P},\mbox{[MASK]}]
\end{equation}

The label ground truth of the very polarity judgement prompt is the same as the corresponding polarity of the original source triplet. 
For negative pairs, because of the mismatch of aspect and opinion terms, judging the polarity of such a pair is meaningless. Thus, in this case, only the aspect-opinion combination part will be fed into the model. 
The whole template $T$ for prompt is like:
\begin{equation}
    T = [P_{1:l_1},A,P_{l_1+1:l_2},O,\mbox{[MASK]}]~T'
\end{equation}
In this way, we can formulate a simple classification task for \Ours~MLM through a prompt $X^p$ according to the input sentence $X$ with the extra polarity judgement prompt $X'^{p}$ as the suffix. Take a manually designed one as an example:
\begin{eqnarray}
    X^p &=& \mbox{The $A$ is $O$? [MASK].} X'^{p}\\
    X'^{p} &=& \mbox{This is [MASK]}
\end{eqnarray}
We use this as an auxiliary task for the encoder part to decide whether to fill in ``yes'' (consistent with ground truth) or ``no'' (inconsistent) for the masked position. In addition, if the polarity judgement prompt is attached, let the encoder part decide the right polarity class to fill in that masked position. 
Let $y^p \in \mathcal{S}=\{\mbox{``yes''},\mbox{``no''},\mbox{POS},\mbox{NEG},\mbox{NEU}\}$ be the polarity class token label for [MASK], the probability of predicting consistency is:
\begin{eqnarray}
\begin{aligned}
p(y^p|X,X^p) =& p(\mbox{[MASK]}=y^p|X,X^p) \\
        =& \frac{\mbox{exp}(\mathbf{w}_{y^p} \cdot h_{\mbox{[MASK]}})}{\sum_{y'^p \in \mathcal{S}}{\mbox{exp}( \mathbf{w}_{y'^p} \cdot h_{\mbox{[MASK]}})}}
\end{aligned}
\end{eqnarray}
where $h_{\mbox{[MASK]}}$ is the hidden vector of the [MASK] position and $\mathbf{w}$ means the weight vector before softmax for the word. 
For this sentiment knowledge enhanced prompt MLM task, we compute the cross-entropy loss of the predictions on masked tokens:
\begin{equation}
    \mathcal{L}_{\mbox{prompt}} =  -\sum \sum y^p \log(p(y^p|X, X^p))
\end{equation}
\subsubsection{Prompt Encoder}
For better prompt representations, we replace the embeddings of pseudo prompt tokens $P_k$ from BART with their differential output embeddings through a prompt encoder and optimize them for the downstream task. 
To associate these pseudo prompt tokens with each other, we choose a bidirectional long-short term memory networks to connect these separate token embeddings and two layers of multi-layer perceptron activated by ReLU to get the processed hidden states.
The prompt encoder takes the embeddings of $l_P$ pseudo prompt tokens $P_k$ from a prompt-token embedding layer $\mathbf{e}_p$ as input and outputs the optimized embedding $h'$. 
\begin{equation}
\begin{aligned}
    E^p =& {\mathbf{e}}_p(P) \\
        =& [{\mathbf{e}}_p(P_1),{\mathbf{e}}_p(P_2)......,{\mathbf{e}}_p(P_{l_P})] \\
        =& [h_1, h_2, ......h_{l_P}]\\
\end{aligned}
\end{equation}
\begin{equation}
    h_k' = \mbox{MLP}([\overrightarrow{\mbox{LSTM}}(h_{0:k-1});\overleftarrow{\mbox{LSTM}}(h_{k:l_P})])
\end{equation}
where $h_k' \in \mathcal{R}^{d}$. The optimized embedding at each position of pseudo prompt tokens will replace the embedding from the layer $\mathbf{e}$ of BART in \Ours~MLM. So the tokens in $T$ will be mapped as:
\begin{equation}
\begin{aligned}
    & [h'_{1:l_1}, {\mathbf{e}}(A), h'_{l_1+1:l_2}, {\mathbf{e}}(O), {\mathbf{e}}(\mbox{[MASK]}) ] \\ 
    & [h'_{l_2+1:l_P},{\mathbf{e}}(\mbox{[MASK]})]
\end{aligned}
\end{equation}
and then be fed into the encoder part while tokens of $A$, $O$ and [MASK] will still be embedded by the BART embedding layer $\mathbf{e}$.
\subsubsection{Prompt Optimization for ABSA}
Given a PLM $\mathcal{M}$ and an input sequence $X$, the pre-trained embedding layer $\mathbf{e} \in \mathcal{M}$ will map every token $[{x_i}_{i=1}^n]$ of $X$ to input embeddings $[\mathbf{e}({x_i}_{i=1}^n)]$. Let $\mathcal{V}_\mathcal{M}$ be the vocabulary of $\mathcal{M}$, the prompt tokens $P_k$ in $T$ and $T'$ satisfy that $P_k \in \mathcal{V}_\mathcal{M}$ as well as the label words $\mathcal{S}$ for [MASK].
As $h'$ is from prompt encoders with trainable parameters, this allows better continuous prompts beyond the original vocabulary $\mathcal{V}_\mathcal{M}$ could express. With the downstream loss, these sentiment knowledge enhanced prompts can be differentially optimized by
\begin{equation}
    \hat{h'}_{1:l_P} = \operatorname*{argmin}_{h'}\mathcal{L}(\mathcal{M}(X,X^p,Y,Y^p))
\end{equation}

\subsection{Generation Framework for ABSA}
For the ABSA task, the model takes a sequence of tokens $X$ as input and hope to generate the target sequence $Y$ as defined above. The input and output sequence starts and ends with special tokens: \textless{}$ s$\textgreater{} and \textless{}$\backslash{}s$\textgreater{}. 
These special tokens are to be generated in $Y$ but we ignore them in equations for simplicity. Given a sequence of tokens $X$:
\begin{equation}
    P(Y|X) = \prod_{t=1}^m p(y_t|X, y_1, y_2,...,y_{t-1})
\end{equation}
To model the index probability $p_t$ for every $t$, we use BART model with pointer network, which roughly consists of an \textbf{Encoder} part and a \textbf{Decoder} part.
\begin{figure}[htb]
    \centering
    \includegraphics[width=0.48\textwidth,]{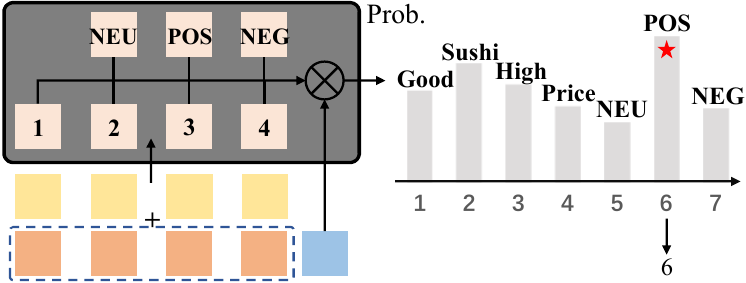}
    \caption{The index generator in the decoder part. The light and dark yellow boxes are token embeddings and outputs from BART Encoder and blue boxes are from BART Decoder as shown in Figure \ref{fig:model}. 
    }
    \label{fig:model_pointer}
\end{figure}
\subsubsection{Encoder}
The encoder part is to encode $X$ into the hidden representation space as a vector $H^e$. 
\begin{eqnarray}
    E^e &=& {\mathbf{e}}(X) = [\mathbf{e}(x_1),\mathbf{e}(x_2)...\mathbf{e}(x_n)] \\
    H^e &=& \mbox{BARTEncoder}(E^e)
\end{eqnarray}
where $E^e, H^e \in \mathcal{R}^{n \times d}$, $d$ is the hidden state dimension and ${\mathbf{e}}$ is the embedding layer.

\subsubsection{Decoder}
The decoder part takes the encoder outputs $H^e$ and previous decoder outputs $y_1, y_2,...,y_{t-1}$ as inputs and decode $y_t$ out. 
As $[{y_j}_{j=1}^{t-1}]$ are token indexes, an Index-to-Token Converter is applied to make a conversion.
\begin{equation}
\tilde{y_j}= \left\{ 
\begin{array}{ll}
X_{y_j}, & \mbox{if\: $y_j$\: is\: a\: pointer\: index}\\
C_{y_j-n}, & \mbox{if\: $y_j$\: is\: a\: polarity\: class\: index}

\end{array}
\right.
\end{equation}

where $C=[c_1, c_2,....c_l]$ is the polarity class token list\footnote{The polarity class token always starts after the pointer indexes of the given sequence, at $n + 1$.}.
After this, we use a BART decoder to encode tokens to get the decoder hidden state for $y_t$.
\begin{eqnarray}
    h_t &=& \mbox{BARTDecoder}(H^e;[\tilde{{y_j}}_{j=1}^{t-1}])
\end{eqnarray}
where $h_t \in \mathcal{R}^{d}$, and the probability distribution $p_t$ of token $y_t$ can be computed as:
\begin{eqnarray}
    \tilde{H^e} &=& \alpha \cdot \mbox{MLP}(H^e) + (1-\alpha)\cdot E^e \\
    C^e &=& {\mathbf{e}}(C) \\
    p_t &=& \mbox{Softmax}([\tilde{H^e};C^e]\cdot h_t)
\end{eqnarray}
where $\tilde{H^e} \in \mathcal{R}^{n \times d}$, $C_e \in \mathcal{R}^{l \times d}$ and $p_t \in \mathcal{R}^{(n+l)} $ is the predicted probability distribution of $y_t$ on all candidate indexes.

\subsection{Training and Inference}
During training, we compute the weighted sum of the cross-entropy losses from sentiment knowledge enhanced prompt-tuning and index predictions for optimization of the generation framework and continuous prompt representations.
\begin{eqnarray}
    \mathcal{L}(\theta) &=& \alpha_1 \cdot \mathcal{L}_{\mbox{prompt}} + \alpha_2 \cdot \mathcal{L}_{\mbox{gen}}
\end{eqnarray}
During inference, we only use $X$ for generation.
We apply beam search to generate the prediction sequence in an auto-regressive way. Then, a specific decoding process is applied to convert the output sequence into required terms and polarities. The corresponding decoding algorithm is included in Appendix G.

\section{Experiments}
In this section, we first introduce the datasets on which the experiments are conducted and detail those baselines to be compared with, then present our main results. 
\subsection{Datasets}
We evaluate our method on three versions of four ABSA datasets.
The first version ($\mathcal{D}_{20a}$) is from \citet{Peng_2020}. They get the pairly annotated opinions from \citet{Wang2017CoupledMA}, the corresponding aspects annotation from \citet{fan-etal-2019-target}, and refine the data into \textless{}$a$, $o$, $s$\textgreater{} triplet form. The second version  $\mathcal{D}_{20b}$~\citep{xu-etal-2020-position} is a revised version of $\mathcal{D}_{20a}$. The missing triplets with overlapping opinions in \citet{Peng_2020} are added. However, from our observation of four datasets in $\mathcal{D}_{20b}$, there still exist quite a few unlabeled triplets, mistakes about polarities and aspect-opinion mismatching. Thus we make some corrections to these datasets and produce our version of them, namely $\mathcal{D}_{21}$. 

We fix up these datasets for the following problems:
\begin{itemize}
    \item Few long phrases are included in opinions in previous annotations whereas they are notably important for classifying sentiments. We add some phrases such as verb-object and adjective-to-verb phrases into annotations to make the corresponding opinion terms more reasonable and complete for their resulting sentiments.
    \item Whether the adverbs of the corresponding adjective or verb in the opinion are included is ambiguous in previous annotations. We uniformly include adverbs in our annotations.
    \item Some opinion terms are actually towards the pronoun but got wrongly annotated to a nearby aspect.
\end{itemize}
Some revised examples and statistical comparisons between $\mathcal{D}_{20a}$, $\mathcal{D}_{20b}$, and $\mathcal{D}_{21}$ are included in Appendix D and E. We will release $\mathcal{D}_{21}$ later.

\subsection{Baselines}
To put our results in perspective, we summarize state-of-the-art models on three ABSA subtasks and compare \Ours~with them. Detailed comparisons of baselines are shown in Appendix F. 
Those three with the suffix ``+'' denote the variant version by \citet{Peng_2020} for being capable of doing \emph{AESC}, \emph{Pair}, and \emph{Triplet}.
They are modified by adding an MLP to determine whether a triplet is correct in the matching stage.
The baseline results marked with $\dagger$ are from \citet{xu-etal-2020-position} and $\sharp$ are from \citet{yan2021unified}. 
The ``PE'' in Table \ref{tb:penga},\ref{tb:pengb},\ref{tb:lcx} indicates the prompt encoder.
More details about implementations and hyper-parameter settings are shown in Appendix A.

\subsection{Main Results}
\begin{table*}[!htb]
\centering
\small

\begin{tabular}{cccc|ccc|ccc|ccc}

\hline
\multirow{2}{*}{Model} & & 14res & & & 14lap & & & 15res & & & 16res & \\
\cline{2-13}
& \emph{AESC} & \emph{Pair} & \emph{Triple}
& \emph{AESC} & \emph{Pair} & \emph{Triple}
& \emph{AESC} & \emph{Pair} & \emph{Triple}
& \emph{AESC} & \emph{Pair} & \emph{Triple} \\
\hline
CMLA+ $\dagger$ & 70.62 & 48.95 & 43.12 & 56.90 & 44.10 & 32.90 & 53.60 & 44.60 & 35.90 & 61.20 & 50.00 & 41.60 \\
RINANTE+ $\dagger$ & 48.15 & 46.29 & 34.03 & 36.70 & 29.70 & 20.0 & 41.30 & 35.40 & 28.0 & 42.10 & 30.70 & 23.30 \\
Li-unified+ $\dagger$ & 73.79 & 55.34 & 51.68 & 63.38 & 52.56 & 42.47 & 64.95 & 56.85 & 46.69 & 70.20 & 53.75 & 44.51 \\
Peng-two-stage $\dagger$ & 74.19 & 56.10 & 51.89 & 62.34 & 53.85 & 43.50 & 65.79 & 56.23 & 46.79 & 71.73 & 60.04 & 53.62 \\
JET-BERT $\sharp$ & - & - & 63.92 & - & - & 50.0 & - & - & 54.67 & - & - & 62.98 \\
OTE-MTL $\sharp$& - & - & 45.05 & - & - & 59.67 & - & - & 48.97 & - & - & 55.83 \\
Dual-MRC $\sharp$ & 76.57 & 74.93 & 70.32 & 64.59 & 63.37 & 55.58 & 65.14 & 64.97 & 57.21 & 70.84 & 75.71 & 67.40 \\
SPAN-BART $\sharp$ & 78.47 & 77.68 & 72.46 & 68.17 & 66.11 & 57.59 & 69.95 & 67.98 & 60.11 & 75.69 & 77.38 & 69.98 \\
\hline
\Ours & \textbf{81.09} & \textbf{78.96} & \textbf{75.01} & \textbf{70.79} & 68.05 & \textbf{60.41} & \textbf{74.20} & \textbf{70.36} & \textbf{64.50} & \textbf{79.81} & \textbf{80.13} & \textbf{74.66} \\
\Ours~w/o PE & 79.44 & 78.30 & 73.70 & 68.10 & \textbf{68.75} & 58.75 & 70.63 & 67.62 & 61.63 & 77.59 & 78.21 & 72.69 \\

\hline
\end{tabular}
\caption{Comparison of F1 scores for \emph{AESC}, \emph{Pair}, and \emph{Triplet} on the  $\mathcal{D}_{20a}$. We highlight the best results in bold.}
\label{tb:penga}
\end{table*}

\begin{table*}[!htb]
\centering
\small
\begin{tabular}{cccc|ccc|ccc|ccc}
\hline
\multirow{2}{*}{Model} & & 14res & & & 14lap & & & 15res & & & 16res & \\
\cline{2-13}
& \emph{P} & \emph{R} & \emph{F1}
& \emph{P} & \emph{R} & \emph{F1}
& \emph{P} & \emph{R} & \emph{F1}
& \emph{P} & \emph{R} & \emph{F1} \\
\hline
CMLA+ $\dagger$ & 39.18 & 47.13 & 42.79 & 30.09 & 36.92 & 33.16 & 34.56 & 39.84 & 37.01 & 41.34 & 42.1 & 41.72 \\
RINANTE+ $\dagger$ & 31.42 & 39.38 & 34.95 & 21.71 & 18.66 & 20.07 & 29.88 & 30.06 & 29.97 & 25.68 & 22.3 & 23.87 \\
Li-unified+ $\dagger$ & 41.04 & 67.35 & 51.0 & 40.56 & 44.28 & 42.34 & 44.72 & 51.39 & 47.82 & 37.33 & 54.51 & 44.31 \\
Peng-two-stage $\dagger$ & 43.24 & 63.66 & 51.46 & 37.38 & 50.38 & 42.87 & 48.07 & 57.51 & 52.32 & 46.96 & 64.24 & 54.21 \\
JET-BERT $\sharp$ & 70.56 & 55.94 & 62.40 & 55.39 & 47.33 & 51.04 & \textbf{64.45} & 51.96 & 57.53 & \textbf{70.42} & 58.37 & 63.83 \\
GTS-BERT $\sharp$ & 67.76 & 67.29 & 67.50 & 57.82 & 51.32 & 54.36 & 62.59 & 57.94 & 60.15 & 66.08 & 69.91 & 67.93 \\
SPAN-BART $\sharp$ & 65.52 & 64.99 & 65.25 & 61.41 & 56.19 & 58.69 & 59.14 & 59.38 & 59.26 & 66.60 & 68.68 & 67.62 \\
\hline
\Ours & 72.79 & \textbf{72.94} & \textbf{72.86} & \textbf{63.40} & \textbf{58.60} & \textbf{60.90} & 62.97 & \textbf{62.06} & \textbf{62.51} & 70.20 & \textbf{73.35} & \textbf{71.74}\\
\Ours~w/o PE & \textbf{73.65} & 67.20 & 70.28 & 60.42 & 57.86 & 59.11 & 60.61 & 61.24 & 60.92 & 67.90 & 71.21 & 69.52 \\
\hline
\end{tabular}
\caption{Comparison of F1, Precision and Recall for \emph{Triplet} on the $\mathcal{D}_{20b}$. We highlight the best results in bold.}
\label{tb:pengb}
\end{table*}

\begin{table*}[!htb]
\centering
\small

\begin{tabular}{cccc|ccc|ccc|ccc}

\hline
\multirow{2}{*}{Model} & & 14res & & & 14lap & & & 15res & & & 16res & \\
\cline{2-13}
& \emph{P} & \emph{R} & \emph{F1}
& \emph{P} & \emph{R} & \emph{F1}
& \emph{P} & \emph{R} & \emph{F1}
& \emph{P} & \emph{R} & \emph{F1} \\
\hline
OTE-MTL & 50.81 & 58.24 & 54.27 & 47.97 & 39.68 & 43.44 & 56.90 & 38.67 & 46.05 & 56.25 & 53.73 & 54.96 \\
JET-BERT & \textbf{72.79} & 30.01 & 42.50 & \textbf{61.80} & 25.40 & 36.00 & \textbf{69.56} & 37.13 & 48.42 & 68.18 & 29.30 & 40.98 \\
GTS-BERT & 61.09 & 60.62 & 60.85 & 56.48 & 49.82 & 52.94 & 58.39 & 55.08 & 56.68 & 63.78 & 61.75 & 62.75 \\
SPAN-BART & 64.36 & 
\textbf{64.24} & 64.30 & 60.41 & 51.95 & 55.86 & 54.00 & 54.10 & 54.05 & 65.19 & 68.47 & 66.79 \\
\hline
\Ours & 66.10 & 63.37 & 64.71 & 61.30 & \textbf{55.32} & \textbf{58.15} & 61.81 & \textbf{62.06} & \textbf{61.93} & \textbf{68.66} & \textbf{69.04} & \textbf{68.85}\\
\Ours~w/o PE & 65.65 & 63.95 & \textbf{64.79} &59.88 & 53.72 & 56.64 & 55.84 & 58.79 & 57.28 & 66.79 & 67.91 & 67.35 \\

\hline
\end{tabular}
\caption{Comparison of F1, Precision and Recall for \emph{Triplet} on $\mathcal{D}_{21}$. We rerun the baseline models according to their open sources on $\mathcal{D}_{21}$ and report their results. The best results are highlighted in bold.}
\label{tb:lcx}
\end{table*}

On $\mathcal{D}_{20a}$, we compare our method on \emph{AESC}, \emph{Pair} and \emph{Triplet}. Experimental results on $\mathcal{D}_{20a}$ are shown in Table \ref{tb:penga}. Our method obtains convincing improvements on all subtasks of all datasets, 3.38\% F1 for \emph{AESC}, 2.09\% F1 for \emph{Pair} and 3.60\% F1 for \emph{Triplet} on average.
Experimental results for \emph{Triplet} on $\mathcal{D}_{20b}$ are shown in Table\ref{tb:pengb}. According to the table, our method achieves better performance than baselines on all datasets. Our method surpasses strong baselines by 7.61 \% at most, with an average of 3.44\% F1 for \emph{Triplet} on all datasets. 
On the coming up $\mathcal{D}_{21}$, we compare our method on \emph{Triplet} in Table \ref{tb:lcx}. \Ours~outperforms representative baselines by an average of 2.30\% F1 on all datasets.
However, compared to the results on $\mathcal{D}_{20b}$, the best performances go down by 8.07\%, 3.78\%, 5.23\%, 2.99\% F1 on 14res, 14lap, 15res, and 16res respectively.
As is reflected by the significant drops of performances, our revised annotations are more complicated and complete thus set a higher demand for the capability of models.
Additionally, \Ours~generally outperforms the model without prompt encoder, except for 14res of $\mathcal{D}_{21}$.
In conclusion, main experimental results demonstrate that \Ours~is a new state-of-the-art method for the ABSA task.

\begin{figure}[htb]
    \centering
    \includegraphics[width=0.49\textwidth]{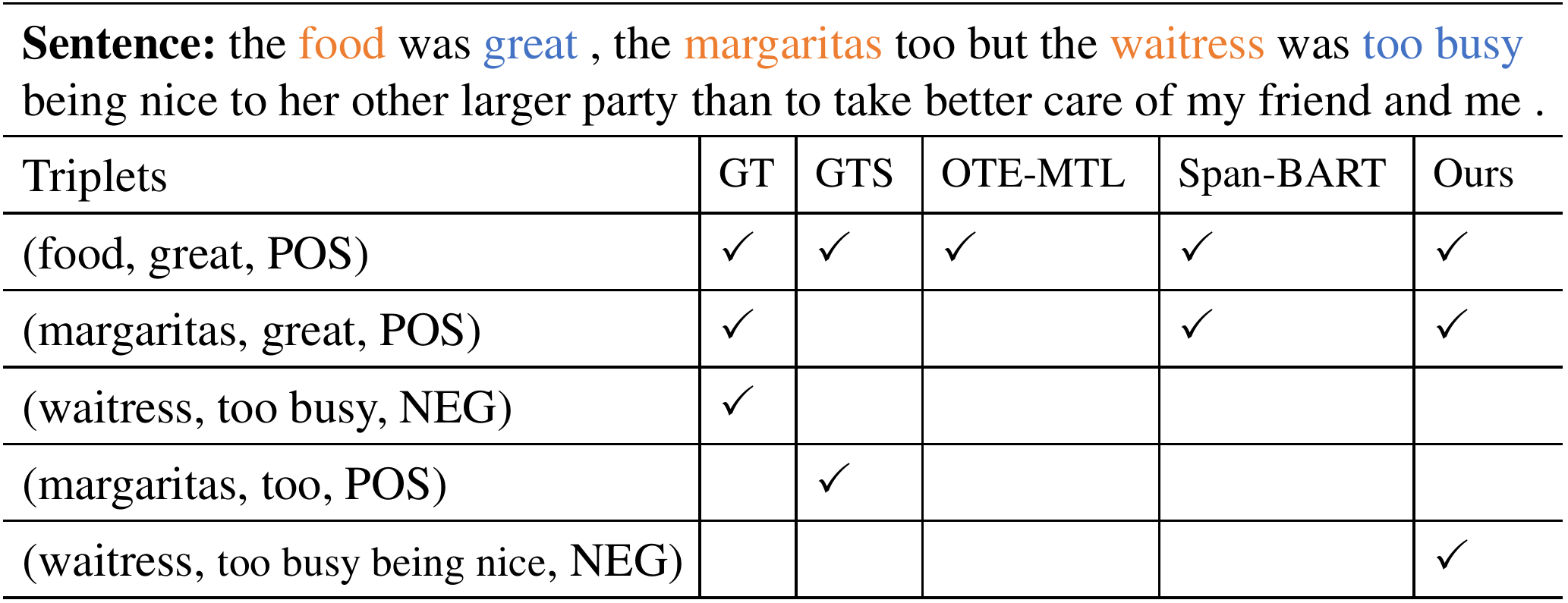}
    \caption{A case study on an input with multiple triplets. GT is the ground truth. 
    }
    \label{fig:case_study}
\end{figure}

To illustrate the differences between models, some triplet prediction cases from some baselines and our \Ours~on a sample sentence are shown in Figure \ref{fig:case_study}. 
For the given example, OTE-MTL could only make one triplet prediction out of three. GTS simply takes the token ``too'' following ``margaritas'' as its opinion term. SPAN-BART is able to give the right predictions on ``food'' and ``margaritas'' as well as our method but fails to identify ``waitress''. Our method gives a larger span on the opinion for ``waitress'', which is comparatively reasonable. Consider that the annotation span for this is actually not that complete. Generally speaking, \Ours~generates more reasonable and accurate outputs in this case. 
These indicate that our method is more helpful to extract correct terms and capture accurate relations between them, which benefit the polarity classification as well. These improvements surely lead to better model performance.
Yet, while our method achieves significant improvements on 14res, 15res and 16res, the improvement on 14lap is relatively marginal.

\subsection{Analysis}
Tuning the LM in unified frameworks using prompting methods with sentiment knowledge shows exciting potentials, but also has its limitations.
We conduct some extensional experiments to further discuss \Ours.
\subsubsection{Few-shot ABSA}
To further evaluate the potential of our method, we conduct few-shot setting experiments with 10\% of training data on four datasets. Due to the instability of tuning on small datasets, we measure average performance across 5 different randomly sampled $\mathcal{D}_{train}$ and $\mathcal{D}_{dev}$, using a fixed set of seeds $\mathcal{D}_{seed}$\footnote{We randomly generate 5 numbers in the range of (0, 6000) from the random function. Here $\mathcal{D}_{seed}$=\{544, 3210, 8, 5678, 744\}}. This gives a more robust measure of model performance and a better estimate of the variance. As previous works about these ABSA subtasks in few-shot setting are scarce, we still adopt some baselines mentioned above for comparison in few-shot setting experiments.
The comparison of experimental results on $\mathcal{D}_{20b}$ and $\mathcal{D}_{21}$ are in Appendix B. These figures show that \Ours~achieves better performances than all baselines on 14lap and 14res and outperforms the majority class of baselines on 15res and 16res. However, our results suffer from high variance. In worst cases, the performance drops up to 10\% F1 below the average. 
In addition, unlike some other prompt-tuning methods, \Ours~won't perform well if most parameters of the LM are frozen, which increases the computational cost of tuning in this process.

\subsubsection{Prompt Length Analysis}
Additionally, to investigate the effect of prompts with different lengths to the ABSA task, we conduct experiments with prompts of various lengths on $\mathcal{D}_{21}$.
For automatically optimized prompts, we simplify the template for this type and control the length by the numbers of pseudo prompt tokens at three fixed positions of the template. Let $N=l_1=l_2-l_1=l_P-l_2$, pseudo prompt tokens at these positions are of the same number $N$. We compare prompts with $N=1,2,3$. Manually designed ones are as shown in Table \ref{tab:prompt_example}.
The comparison figure is included in Appendix C. In most cases, continuously optimized prompts achieve comparable or higher performance than manually designed ones. 
According to the figure, it is clear that the number of pseudo prompt tokens in a template have notable influence on model performance as subtasks and datasets change. As more pseudo tokens provide more trainable parameters, they enhance the capability of the model yet require more data and time to be properly trained.

\subsubsection{Multi-triplet Analysis}
Besides, we compare the model performance of SPAN-BART and \Ours~on $\mathcal{D}_{21}$ for multi-triplet cases, with more than one aspect or opinion within the input. These data account for about 70\% of the total.
As shown in Table \ref{tb:multi-ins}, \Ours~improves the Precision quite a lot while keeps the Recall changing little, resulting in the general lift on F1. 
Consider the large proportion of multi-triplet cases in data, the improvements in main results can be largely attributed to better performance on them.
With the LM capturing relations between multiple aspects and opinions better, \Ours~performs better on multi-triplet cases and thus benefits the overall performance.

\begin{table}[!htb]
\centering
\small

\begin{tabular}{clcccc}
\hline
\multirow{2}{*}{Model} & \multirow{2}{*}{Metric} & \multicolumn{4}{c}{Multi-Triplet} \\
\cline{3-6}
& & 14res & 14lap & 15res & 16res \\
\hline
\multirow{3}{*}{SPAN-BART} & \emph{P} & 66.27 & 63.07 & 58.29 & 59.14\\
& \emph{R} & 59.35 & 48.76 & 48.05 & 71.08 \\
& \emph{F1} & 62.62 & 55.00 & 52.68 & 64.56 \\
\multirow{3}{*}{\Ours} & \emph{P} & 70.13 & 68.09 & 64.48 & 77.03 \\
& \emph{R} & 57.93 & 49.48 & 52.35 & 61.76 \\
& \emph{F1} & 63.45 & 57.31 & 57.79 & 68.55 \\
\hline
\end{tabular}
\caption{Extensional comparison of results for multi-triplet cases on $\mathcal{D}_{21}$ dataset.}
\label{tb:multi-ins}
\end{table}

\subsubsection{Invalid Prediction Analysis}
Since those subtasks require formatted sequence output to decode as described in Section 3, we estimate the invalid prediction rates for \emph{Triplet} of the generative framework SPAN-BART and \Ours, on error length~(Err-length) and error order~(Err-order), respectively. A valid triplet prediction $p_i$ should contain five indexes. The start index should be smaller or equal to the paired end index. From Table \ref{tb:error}, \Ours~could achieve sufficiently low invalid rates for both error cases on most datasets. 
lower invalid rate on 14res, 15res and error length decrease much on 16res.
Lower invalid rates compared to SPAN-BART demonstrates that a better way of introducing knowledge for unified frameworks leads to well-formatted index sequence predictions, which lay a solid foundation for better actual performance in practical scenarios.

\begin{table}[!htb]
\centering
\small

\begin{tabular}{clcccc}

\hline
Model & Error(\%) & 14res & 14lap & 15res & 16res  \\
\hline
\multirow{2}{*}{SPAN-BART} & Err-length & 0.48  & 0.77 & 1.41 & 1.40 \\
              & Err-order & 1.75 & 3.70 & 3.26 & 3.26 \\
\hline
\multirow{2}{*}{\Ours} & Err-length & 0.41 & 1.01 & 1.24 & 0.31 \\
              & Err-order & 1.63 & 3.68 & 3.11 & 4.29 \\
\hline
\end{tabular}
\caption{The errors for \emph{Triplet} on the test set of $\mathcal{D}_{20b}$ dataset.}
\label{tb:error}
\end{table}

\section{Conclusion}
In this work, we propose \Ours~to utilize sentiment knowledge enhanced prompts for the ABSA task, which is a task-adaptive modification to a unified generative framework. 
we inject sentiment knowledge regarding aspects, opinions, and polarities from ground truth into prompts and explicitly model term relations via consistency and polarity judgment templates. 
Through \Ours~MLM during training, the encoder part of the framework learn to explicitly model sentiment relations between aspects and opinions. 
Likewise, we introduce sentiment knowledge for accurate term extraction and sentiment classification by this tuning process. 
Additionally, due to mistakes and incompleteness in annotations, we revise four public datasets to form our version $\mathcal{D}_{21}$.
We conduct experiments on three versions of four benchmark datasets. \Ours~achieves significant improvements on ABSA subtasks on all versions of datasets. 


\clearpage
\bibliography{ref}
\clearpage
\def\year{2022}\relax

\section{A. Hyper-parameter settings in sour experiments}
All experiments are completed on a single 32G Nvidia-V100 GPU. It takes about 2 hours to finish 50 epochs.
Our model is constructed with BART-base pre-trained model.
The hyper-parameters to get the best results on every dataset are as in Table \ref{tab:hyper}.
\begin{table}[htb]
    \centering
    \begin{tabular}{c|c|c|c|c}
    \hline
    Dataset & 14lap & 14res & 15res & 16res \\
    \hline
        Training Epochs & \multicolumn{4}{c}{50}\\
    \hline
        Batch Size & \multicolumn{3}{c|}{8} & 32 \\
    \hline
        Learning Rate & \multicolumn{3}{c|}{5e-5} & 1e-4 \\
    \hline
        Hidden Size & \multicolumn{4}{c}{1024}\\
    \hline
    \end{tabular}
    \caption{Hyper-parameters used to get the best performance in training}
    \label{tab:hyper}
\end{table}

For AESC, Pair and Triplet, a prediction result is correct only when all the span boundaries are exactly matched and the predicted sentiment polarity are accurately classified. For $\mathcal{D}_{20a}$, We report F1 scores of AESC, Pair and Triplet tasks for all datasets. For $\mathcal{D}_{20b}$ and $\mathcal{D}_{21}$, we report the precision (P), recall (R), and F1 scores of Triplet task for them.

\section{B. Experimental results under few-shot setting scenario}

\begin{figure}[htb]
    \centering
    \includegraphics[width=0.48\textwidth,]{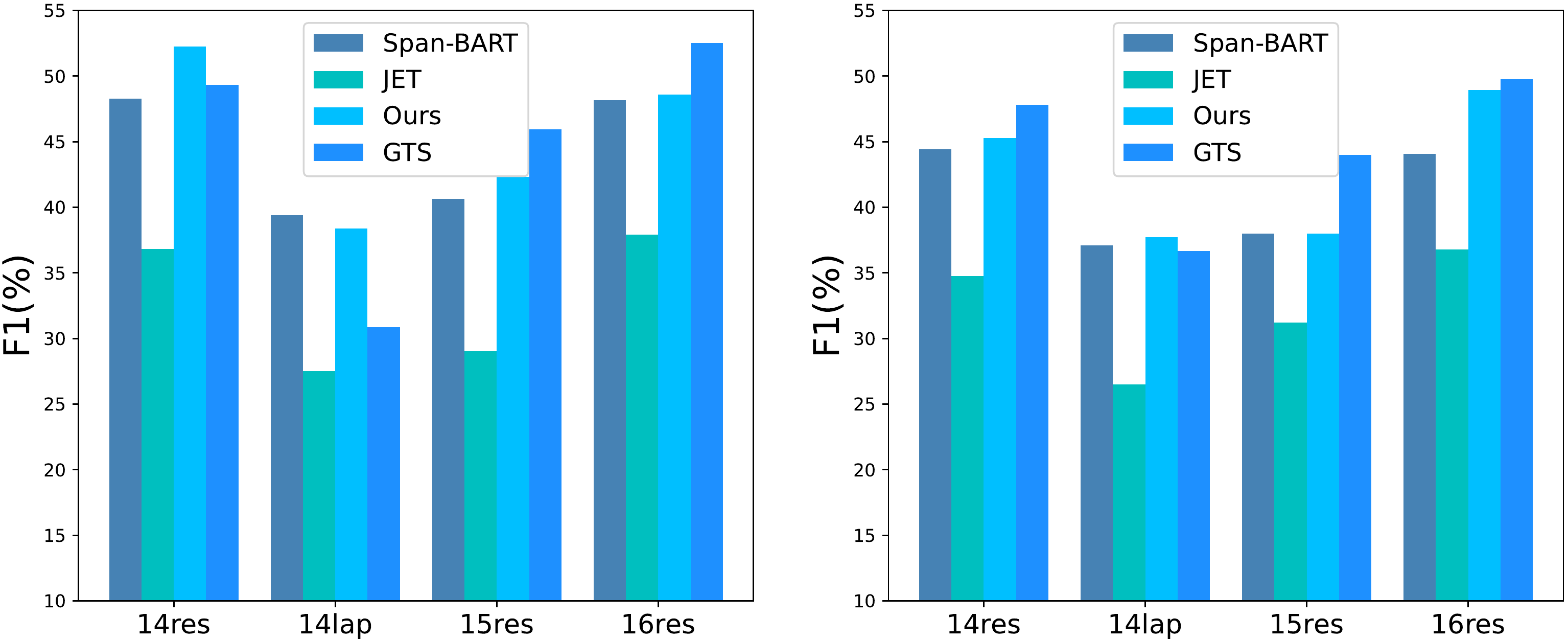}
    \caption{Comparison of F1 scores for \emph{Triplet} on the $\mathcal{D}_{20b}$ and $\mathcal{D}_{21}$ datasets under few-shot setting with 10\% of data.}
    \label{fig:fewshot}
\end{figure}

\section{C. Experimental results for different prompts}
\begin{figure}[htb]
    \centering
    \includegraphics[width=0.45\textwidth,]{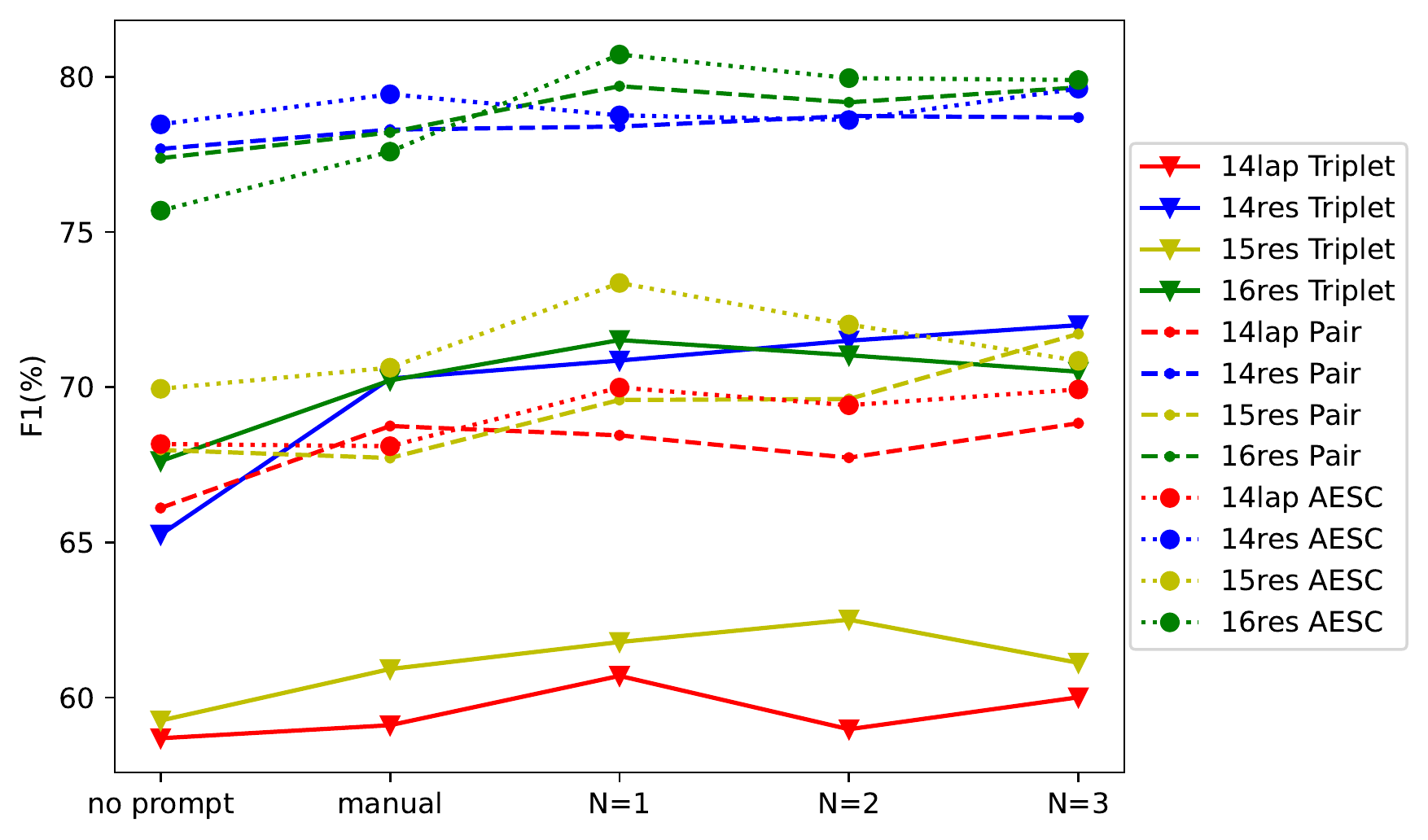}
    \caption{Comparison of F1 scores for \emph{AESC}, \emph{Pair} and \emph{Triplet} on the $\mathcal{D}_{21}$ dataset.}
    \label{fig:prompt}
\end{figure}

We compare prompts with different $N$ and the handcrafted one. The results are shown in Figure \ref{fig:prompt}.

\section{D. Examples of our revision to the annotations in $\mathcal{D}_{20b}$}
Due to the three problems in previous datasets mentioned in the paper, we revise some annotations in $\mathcal{D}_{20b}$ to form our version $\mathcal{D}_{21}$. To get these problems through, we pick some typical annotations in $\mathcal{D}_{20b}$ and our corresponding revised version as examples in Table \ref{tab:revised_example}. 
\begin{table*}[htb]
    \centering
    \begin{tabular}{|c|c|c|c|}
    \hline
       \multirow{5}{*}{Sentence}  & & & \small{What can you say about a place where the}\\
       & The battery lasts as advertised & We ordered a tuna melt-it & \small{waitress brings out the wrong entree, then}\\
       & (give or take 15-20 minutes), & came with out cheese which & \small{verbally assaults your 80 year old grand-}\\
       & and the entire user experience  & just made it a tuna sandwich. & \small{mother and gives her lip about sending it}\\
       & is very elegant.& & \small{back (which she did politely, by the way).} \\
    \hline
        \multirow{3}{*}{In $\mathcal{D}_{20b}$} && (tuna melt, with out, NEG) & \\
        &  (user experience, elegant, POS) & (cheese, with out, NEG) &(entree, wrong, NEU)\\
        & & (tuna sandwich, with out, NEG) &\\
    \hline
        \multirow{2}{*}{In $\mathcal{D}_{21}$} 
        &  & & \small{(waitress, brings out the wrong entree, NEG)}\\
        & \small{(battery, lasts as advertised, POS)} & \small{(tuna melt, with out cheese, NEG)} & (waitress, verbally assaults your \\
        & (user experience, elegant, POS)& & 80 year old grandmother, NEG)\\
        & & & \small{(waitress, gives her lip about sending it, NEG)}\\
    \hline
    \end{tabular}
    \caption{Some examples of our revision to the annotations of samples. In $\mathcal{D}_{20b}$, some aspect-opinion pairs are missing and some are wrongly matched. There also exist some complex cases not properly annotated. While in $\mathcal{D}_{21}$, we correct those mistakes and use long phrases such as verb-object phrase to annotate complex samples. }
    \label{tab:revised_example}
\end{table*}

\section{E. Statistical comparison between three versions of four datasets.}
A detailed statistical comparison between $\mathcal{D}_20{a}$, $\mathcal{D}_20{b}$ and $\mathcal{D}_{21}$ is presented in Table \ref{tab:data}. $\mathcal{D}_{21}$ has a bit more sentences and triplets annotations than $\mathcal{D}_{20b}$ due to our revision. 
\begin{table*}[htb]
  \centering
  \small
  \setlength{\tabcolsep}{4.5pt}
  \begin{tabular}{cc|cc|cc|cc|cc}
    \hline
  \multicolumn{2}{l|}{\multirow{2}{*}{Dataset}} & \multicolumn{2}{c}{14res} & \multicolumn{2}{c}{14lap} & \multicolumn{2}{c}{15res} & \multicolumn{2}{c}{16res}                                                                                               \\
    \cline{3-10}
    \multicolumn{2}{l|}{}                         & $N_s$  & $N_p$  & $N_s$ & $N_p$  & $N_s$  & $N_p$  & $N_s$ & $N_p$                                                                                                                           \\
    \hline
    \multirow{3}{*}{\emph{$\mathcal{D}_{20a}$}} & train        & \small{1300} & \small{2145} & \small{920} & \small{1265} & \small{593} & \small{923}  & \small{842} & \small{1289} \\
                                & dev          & \small{323}  & \small{524}  & \small{228} & \small{337}  & \small{148} & \small{238}  & \small{210} & \small{316}                                                                                                                        \\
                                & test         & \small{496} & \small{862}  & \small{339} & \small{490}  & \small{318} & \small{455}  & \small{320} & \small{465} \\
    \hline
    \multirow{3}{*}{\emph{$\mathcal{D}_{20b}$}}        & train        & \small{1266} & \small{2338} & \small{906} & \small{1460} & \small{605} & \small{1013} & \small{857} & \small{1394} \\
                                & dev          & \small{310}  & \small{577}  & \small{219}  & \small{346}  & \small{148} & \small{249}  & \small{210}  & \small{339}                                                                                                                      \\
                                & test         & \small{492}  & \small{994}  & \small{328} & \small{543} & \small{148} & \small{485}  & \small{326}  & \small{514}                                                                                                                        \\
    \hline
    \multirow{3}{*}{\emph{$\mathcal{D}_{21}$}}        & train    & \small{1266} & \small{2436} & \small{906}  & \small{1465} & \small{605}  & \small{1036} & \small{857} & \small{1459}  \\
                                & dev          & \small{310} & \small{598}  & \small{219} & \small{372}  & \small{148}  & \small{273}  & \small{210} & \small{355}                                                                                                                           \\
                                & test         & \small{492}  & \small{1043} & \small{328}  & \small{567} & \small{148}  & \small{512} & \small{326}  & \small{536}   \\
  \hline
  \end{tabular}
  \caption{The statistics of four datasets, where the $N_s$, $N_p$ denote the numbers of sentences, aspect terms, opinion terms, and the \textless{}$a$, $o$\textgreater{}  pairs, respectively. These three versions of four datasets can be applied to all ABSA subtasks mentioned above, including \emph{AE}, \emph{OE}, \emph{ALSC}, \emph{AOE}, \emph{AESC}, \emph{Pair} and \emph{Triplet}.}
  \label{tab:data}
\end{table*}
\section{F. Comparison between different baselines in our experiments}
To make fair comparison between baselines used in our experiments, we summarize the their task formulations, the backbone in their models, and the subtasks they can do in Table \ref{tab:baseline}.
\begin{table*}[htb]
  \centering
  \small
  \setlength{\tabcolsep}{5pt}
  \begin{tabular}{ccccccccccc}
  \hline
    Baselines            & \multicolumn{1}{c}{E2E} & Task Formulation             & Backbone                & \multicolumn{1}{c}{\emph{AE}} & \multicolumn{1}{c}{\emph{OE}} & \multicolumn{1}{c}{\emph{ALSC}} & \multicolumn{1}{c}{\emph{AOE}} & \multicolumn{1}{c}{\emph{AESC}} & \multicolumn{1}{c}{ \emph{Pair} } & \multicolumn{1}{c}{\emph{Triplet}} \\
    \hline
    RINANTE+             & -                       & Seq.Tagging                 & LSTM+CRF                           & \ding{51}              & \ding{51}              & \ding{51}              & -                       & \ding{51}                & \ding{51}                 & \ding{51}                    \\
    CMLA+                & -                       & Seq.Tagging                 & Attention                         & \ding{51}              & \ding{51}              & \ding{51}              & -                       & \ding{51}                & \ding{51}                 & \ding{51}                    \\
    Li-unified+          & -                       & Seq.Tagging                 & LSTM                               & \ding{51}              & \ding{51}              & \ding{51}              & -                       & \ding{51}                & \ding{51}                 & \ding{51}                    \\
    Peng-two-stage       & -                       & Seq.Tagging                 & LSTM+GCN                           & \ding{51}              & \ding{51}              & \ding{51}              & -                       & \ding{51}                & \ding{51}                 & \ding{51}                    \\
    JET-BERT             & \ding{51}               & Seq.Tagging                 & BERT                             & \ding{51}              & \ding{51}              & \ding{51}              & -                       & \ding{51}                & \ding{51}                 & \ding{51}                    \\
    Dual-MRC             & -                       & Span.MRC                    & BERT                               & \ding{51}              & -                      & \ding{51}              & \ding{51}               & \ding{51}                & \ding{51}                 & \ding{51}                    \\
    OTE-MTL & \ding{51} & Seq.Tagging & LSTM &  - &  - &  - &  -  &  -   &  -    & \ding{51} \\  
    GTS-BERT & \ding{51} & Grid.Tagging & BERT & \ding{51} & \ding{51} & - & \ding{51}  &  \ding{51}  & \ding{51}  & \ding{51} \\  
    JET-BERT & \ding{51} & Seq.Tagging & BERT & \ding{51} & \ding{51} & - & \ding{51}  &  \ding{51}  & \ding{51}  & \ding{51} \\ 
    SPAN-BART & \ding{51} & Span.Generation & BART &  \ding{51} &  \ding{51} &  \ding{51} &  \ding{51}  &  \ding{51}   &  \ding{51}    & \ding{51} \\  
    \hline

    Ours & \ding{51}   & Span.Generation  & BART   &  \ding{51} &  \ding{51} &  \ding{51} &  \ding{51}  &  \ding{51}   &  \ding{51}    & \ding{51}        \\
    \hline
  \end{tabular}
  \caption{The baselines in our experiments. ``E2E'' is short for End-to-End, which means the model should output all the subtasks' results synchronously rather than requiring any preconditions, e.g., pipeline methods. }
  \label{tab:baseline}
\end{table*}
\section{G. The decoding algorithm for converting process}
The decoding algorithm for converting the predicted index sequence to aspect spans, opinion spans and polarities is shown in Algorithm \ref{alg:decode}.
\begin{algorithm}[htb]
  \begin{algorithmic}[1]
    \caption{Decoding Algorithm for the \emph{Triplet}}
    \Require $n$, the number of tokens in $X$; target sequence $Y=[y_1, ..., y_m]$; and we have $y_t \in [1, n+l]$
    \Ensure  Target span set $L=\{(a_1^{start}, a_1^{end}, o_1^{start}, o_1^{end},$ $s_1),...,(a_i^{start}, a_i^{end}, o_i^{start}, o_i^{end}, s_i), ...\}$    

    \State $L=\{\}, e=[], i=1$
        \While{$i<=m$}
          \State $y_i = Y[i]$
          \If{$y_i>n$}
          \State  $L.add((e, C_{y_i-n}))$
          \State $e=[]$
          \Else
          \State $e.append(y_i)$
          \EndIf
          \State $i+=1$
        \EndWhile
    \State \Return{$L$}
  \label{alg:decode}
  \end{algorithmic}
\end{algorithm}

\end{document}